\documentclass{article}
\usepackage{spconf,amsmath,graphicx}
\usepackage{algpseudocode}
\usepackage{algorithm}
\usepackage{diagbox}
\usepackage{multirow}
\usepackage{multicol}

\algnewcommand\algorithmicforeach{\textbf{for each}}
\algdef{S}[FOR]{ForEach}[1]{\algorithmicforeach\ #1\ \algorithmicdo}

\title{A CNN framework based on line annotations\\ for detecting nematodes in microscopic images}
\makeatletter
\def\@name{\textit{Long Chen\rlap{,}\textsuperscript{1}} \textit{Martin Strauch\rlap{,}\textsuperscript{1}} \textit{Matthias Daub\rlap{,}\textsuperscript{2}} \textit{Xiaochen Jiang\rlap{,}\textsuperscript{1}} \textit{Marcus Jansen\rlap{,}\textsuperscript{3}} \\ \textit{Hans-Georg Luigs\rlap{,}\textsuperscript{3}} \textit{Susanne Schultz-Kuhlmann\rlap{,}\textsuperscript{4}} \textit{Stefan Krüssel\rlap{,}\textsuperscript{4}} \textit{Dorit Merhof\textsuperscript{1}}
	\thanks{*This work was funded by the Germany Ministry of Education and Research, project KMU-innovativ-19: PheNeSens, grant number 031B0474C}}
\makeatother

\address{\textsuperscript{1}Imaging and Computer Vision, RWTH Aachen University, Aachen, Germany\\ \textsuperscript{2}Julius K\"uhn Institute: Federal Research Centre for Cultivated Plants, Elsdorf, Germany\\
	\textsuperscript{3}LemnaTec GmbH, Aachen, Germany \\
\textsuperscript{4}Landwirtschaftskammer Niedersachsen - Pflanzenschutzamt (PSA), Hannover, Germany}
\begin{document}
%\ninept
%
\maketitle
\begin{abstract}
Plant parasitic nematodes cause damage to crop plants on a global scale. Robust detection on image data is a prerequisite for monitoring such nematodes, as well as for many biological studies involving the nematode \emph{C. elegans}, a common model organism. Here, we propose a framework for detecting worm-shaped objects in microscopic images that is based on convolutional neural networks (CNNs). We annotate nematodes with curved lines along the body, which is more suitable for worm-shaped objects than bounding boxes. The trained model predicts worm skeletons and body endpoints. The endpoints serve to untangle the skeletons from which segmentation masks are reconstructed by estimating the body width at each location along the skeleton. With light-weight backbone networks, we achieve 75.85\% precision, 73.02\% recall on a potato cyst nematode data set and 84.20\% precision, 85.63\% recall on a public \emph{C. elegans} data set.
\end{abstract}
\begin{keywords}
worm detection, CNN, line annotations
\end{keywords}

\section{Introduction}

\subsection{Motivation}
\label{sec:moti}

Many nematode (roundworm) species are parasitic on crops, such as potatoes, sugar beets and soybeans, causing billions of losses in agriculture worldwide~\cite{nemaCost}. Robust nematode detection in microscopic images is a prerequisite for quantifying nematode infestation based on (soil) samples and for phenotyping, i.e.\ measuring quantitative features that characterize the nematodes. Moreover, the nematode \emph{C. elegans} is an important model organism in biology, having, for example, been used for high-throughput screening of antimicrobial drugs~\cite{antimicrobial}. A lesioning study of \emph{C. elegans} motorneurons served to infer the function of individual neurons~\cite{motorneurons}. Large-scale chemical and RNAi screens using nematodes are also widespread~\cite{rnaScreen}. 

\subsection{Contribution}
We introduce a CNN-based approach for detecting worm-shaped objects in microscopic images (Section~\ref{ssec:skel}). The worms are long and thin, i.e.\ they extend over a large range, but cover only a small number of pixels. Thus, we propose to represent worms by curved lines along the body instead of bounding boxes as they are used by most object detection approaches~\cite{detector}. Given well-estimated worm skeletons and endpoints, overlapping worms can be untangled (Section~\ref{ssec:untange}) and segmentation masks can be reconstructed (Section~\ref{ssec:recons}). 

We evaluate our method on a potato cyst nematode (\emph{Globodera} spp.) data set generated by Pflanzenschutzamt (PSA) Niedersachsen and a public \emph{C. elegans} data set (BBBC010~\cite{antimicrobial}), which are representative of sparse- and dense-object data with dirty and clean background, respectively (Sections~\ref{sec:dataset}-\ref{sec:results}). %Our approach should be directly applicable to other worm detection tasks. 

\subsection{Relationship to prior work}
\label{sec:intro}

Previous work focusses largely on untangling worms on clean backgrounds and on phenotyping. In~\cite{wormToolBox, wormModel, wormModel2}, a graph is constructed from skeleton segments for each worm cluster. Individuals are untangled by searching for best fits of the learned worm model while minimizing overlap. Worm Tracker 2.0~\cite{wormTracker} was developed to record the behavior of worms and can extract 702 features relevant to behavioral phenotypes. 

However, the approaches mentioned above can only be applied to pure samples in which worms are relatively easy to segment (as in~\cite{wormToolBox}). In contrast, microscopic images acquired from soil samples (Section~\ref{sec:dataset}) contain eggs, nematode cyst fragments and other organic debris. Recent advances in deep learning have greatly improved detection performance for objects in complex contexts. Standard detection approaches~\cite{detector} make predictions in the form of bounding boxes, which are a good representation of approximately convex objects, but not very informative for elongated worms. Instead, we use curved lines along the worm body as a more suitable representation.

Worms in the image are likely to overlap, especially in case of a large number of worms per area. Different from the model search approaches used by~\cite{wormToolBox, wormModel, wormModel2}, we find that individuals can be well untangled with simple geometric criteria, when the endpoints (head and tail) are known. Therefore, the network is trained to output both the skeleton and endpoints. To handle the instability of the training due to highly unbalanced positive and negative pixels, we adopt a focal loss~\cite{focalLoss} with reduced penalty around the positive pixels, which is inspired by CornerNet~\cite{cornerNet}. 

After skeletons of individuals are obtained, we reconstruct segmentation masks by estimating the body width at each skeleton pixel. The entire pipeline outputs object segmentations, requiring only skeleton annotations for training.

\section{Our Approach}
\label{sec:approach}

\subsection{Skeleton and endpoint prediction}
\label{ssec:skel}

We employ the standard U-Net architecture~\cite{unet} as the CNN component of our framework. Two branches are added on the last feature map for predicting the worm \emph{skeleton} and body \emph{endpoints}, respectively. Each branch consists of one feature convolutional layer (ReLu activation) with 64 channels and a following output layer with 2 classes (softmax activation). 

A common problem of dense prediction at each pixel with highly unbalanced positive and negative labels is that the model will degenerate to predict all pixels as the majority. To make the training gradually focus on the mispredicted minority, we apply a variant~\cite{cornerNet} of the focal loss~\cite{focalLoss}: 

\begin{equation*}
L=\frac{-1}{NHW}\sum_{i=1}^{H} \sum_{j=1}^{W} 
\begin{cases}
(1-p)^\gamma \log p, & \text{$y_{ij}=1$}\\
(1-w_{ij})^\beta (p)^\gamma \log (1-p), & \text{$y_{ij}=0$ ,}
\end{cases}
\end{equation*} 

\noindent where $H$ and $W$ are the image height and width, $N$ is the number of objects in the image, $p$ is the probability of being a positive label. The focusing parameter $\gamma$~\cite{focalLoss} is set to 2. 

We compute the weight map $w_{ij}$ by applying a 1D unnormalized Gaussian function to the distance transforms~\cite{distTran} of the ground truth. The distance-based weighting reduces the penalty of negative pixels around positive pixels, giving a certain degree of tolerance to the offset in both annotation and training. The standard deviation $\sigma$ of the Gaussian determines the distance within which the penalty is reduced. During the training process, this "ground truth slack" is applied separately to the skeletons and to the endpoints. %Our experiments (Section~\ref{sec:results}) show that it has a significant effect on the performance.

The hyperparameter $\beta$ is set to 4 for all experiments. The total training loss is the sum of the losses of both branches.

\subsection{Worm untangling}
\label{ssec:untange}

\begin{algorithm} \caption{Untangle individual skeletons}
	\label{alg:untangle}
	\begin{algorithmic}[1]
		\Require skeleton $skel$, predicted endpoints $ep$
		\State find endpoints $ep\_geo$, intersections $ip\_geo$ of $skel$
		\ForEach {$p \in \{ ep \} $}
		\If {$ p \> \text{ matches\_one\_of } \{ ep\_geo \} $}
		\State continue
		\ElsIf  {$ p \> \text{ matches\_one\_of } \{ p\_skel \} $}
		\State $\text{cut the skeleton}$
		\EndIf
		\EndFor
		
		\State {recompute $ip\_geo$ of the cut $skel$}
		
		\ForEach {$p \in \{ ip\_geo \} $}
		\State cut the skeleton at intersection $p$
		\State get all cut segments
		\State compute steering angles $sA$ between segment pairs
		\While {\#remaining\_segments $\geq 2$}
		\State connect segments with the smallest $sA$ 
		\EndWhile
		\EndFor
	\end{algorithmic}
\end{algorithm}

Skeletons of overlapping worms can be untangled with the help of the \emph{endpoints} (head/tail; $ep$ for brevity) predicted by the CNN. Based only on the geometry of the skeletons,
points on the skeletons ($p\_skel$) can be classified as \emph{geometric endpoints}, intersections and line points ($ep\_geo$, $ip\_geo$, $lp\_geo$). %In the final results, the predicted endpoints should be a subset of the geometric endpoints. 

\begin{figure}[htb]
	
	\begin{minipage}[b]{.32\linewidth}
		\centering
		\centerline{\includegraphics[width=\linewidth]{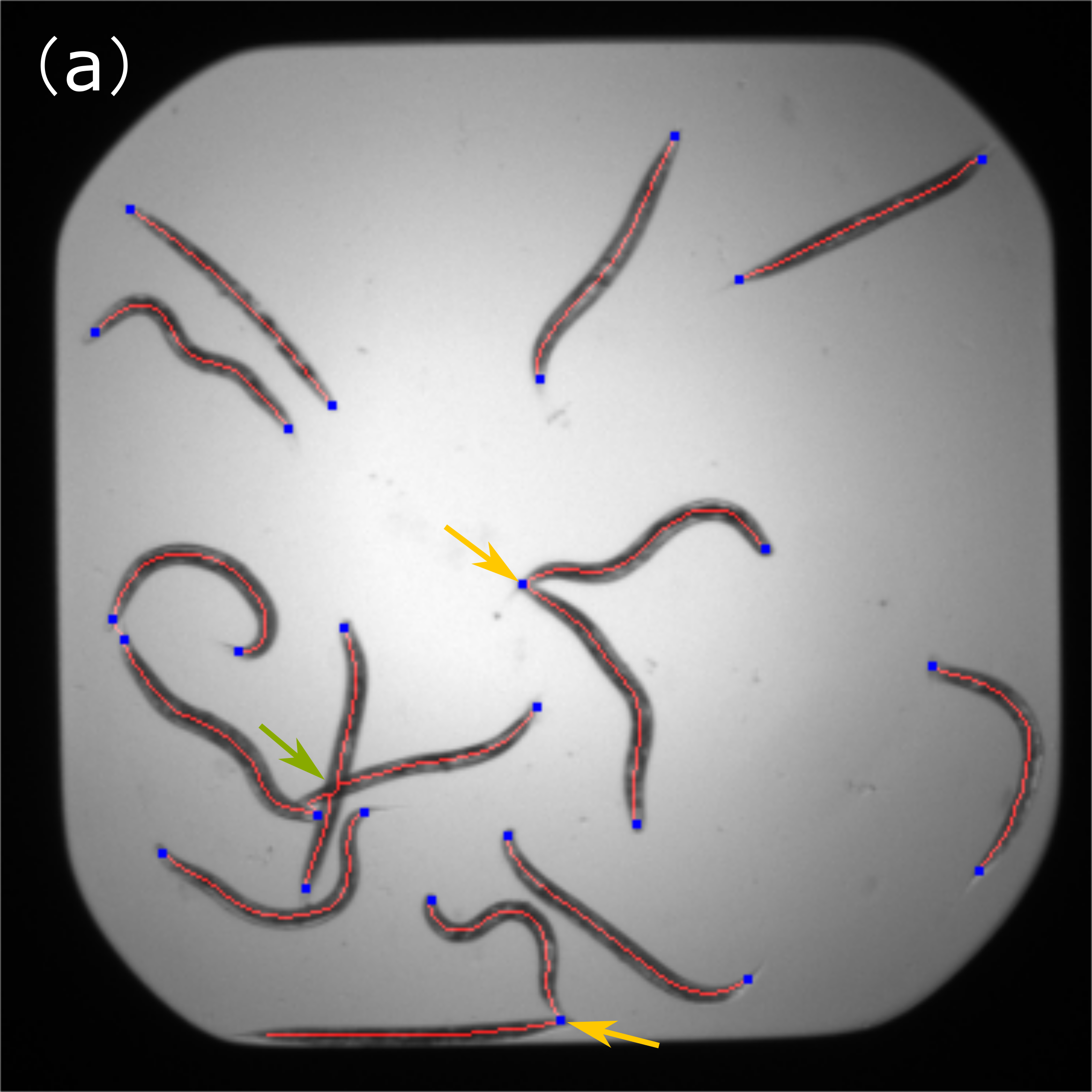}}
	\end{minipage}
	\hfill
	\begin{minipage}[b]{0.32\linewidth}
		\centering
		\centerline{\includegraphics[width=\linewidth]{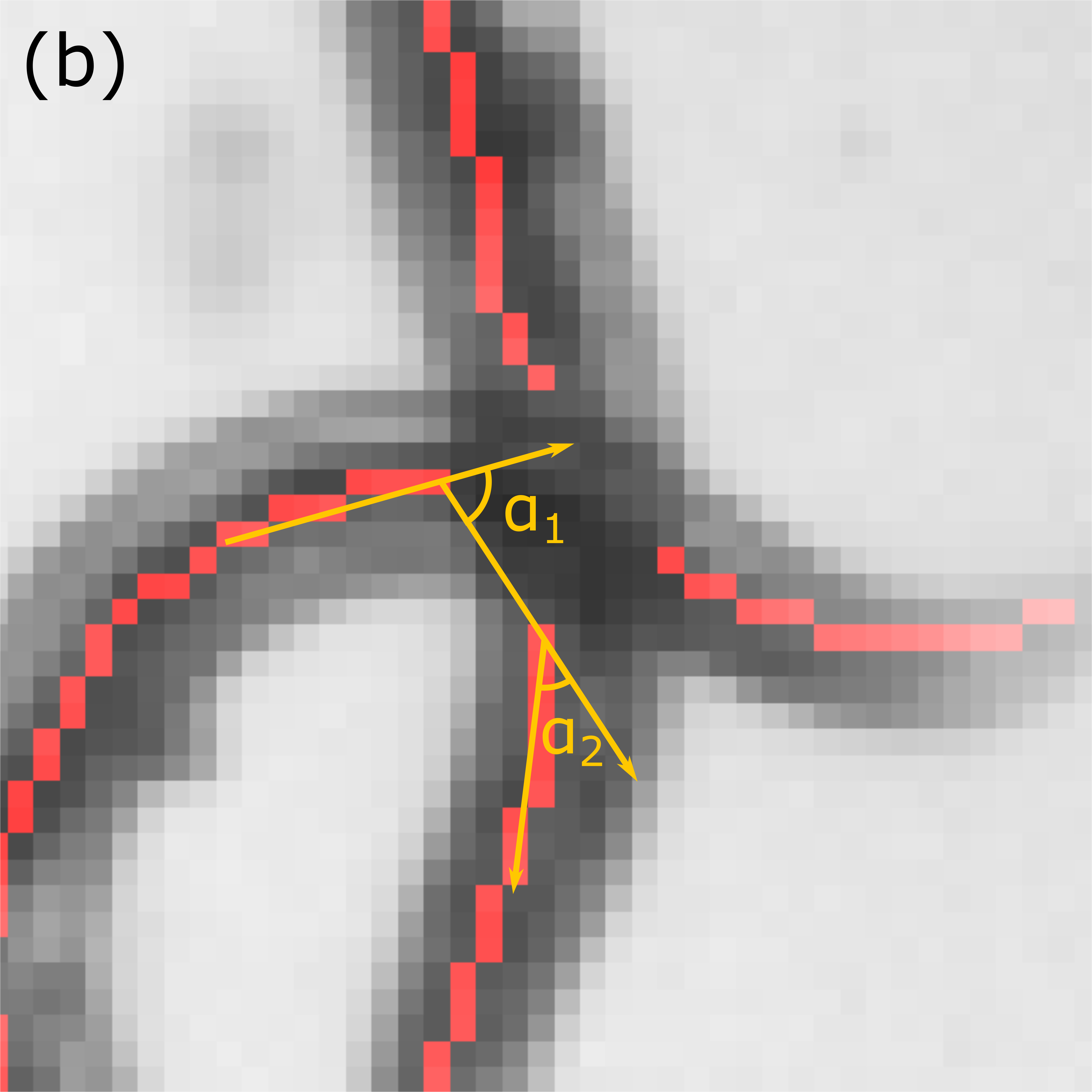}}
	\end{minipage}
	\hfill
	\begin{minipage}[b]{0.32\linewidth}
		\centering
		\centerline{\includegraphics[width=\linewidth]{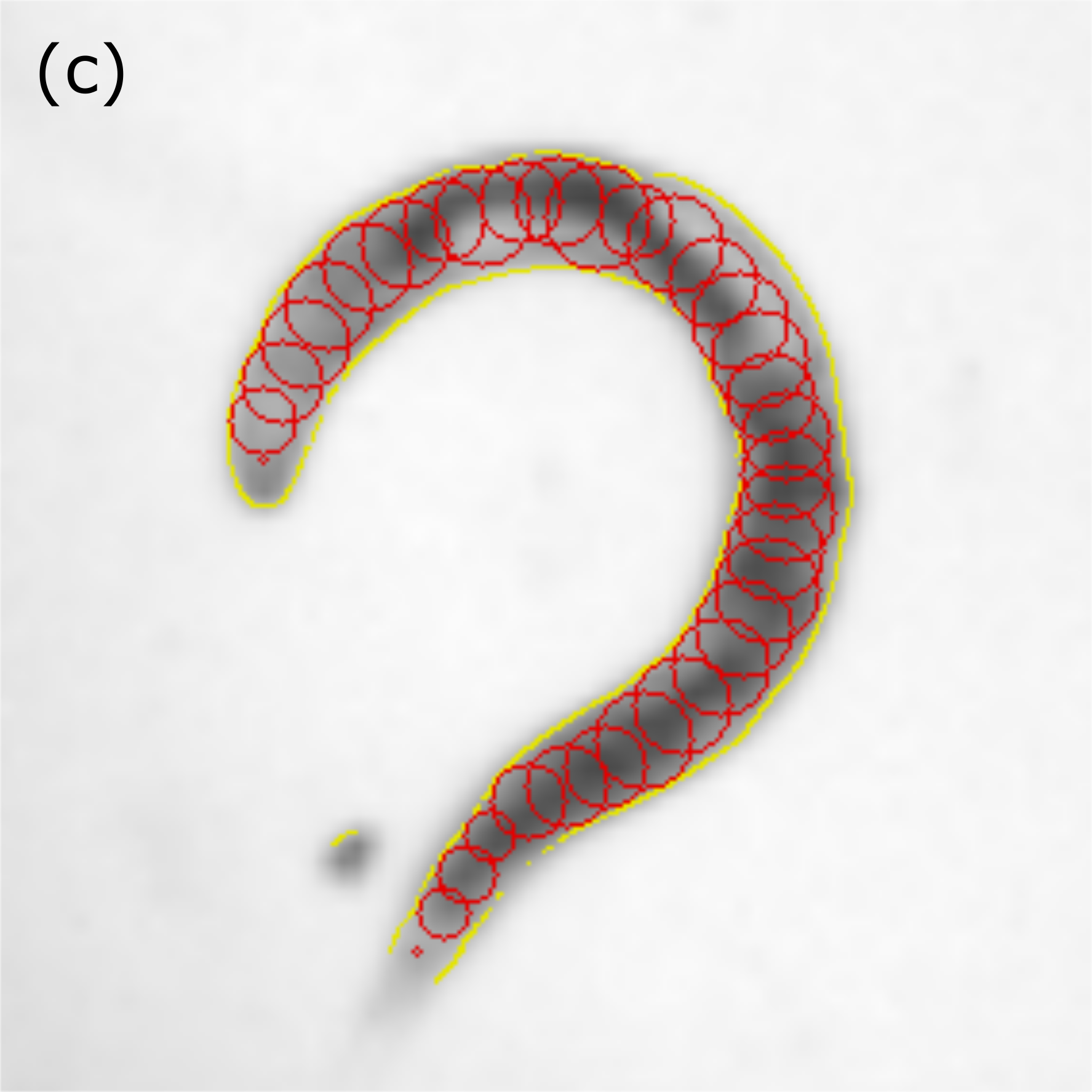}}
	\end{minipage}
	
	\caption{\textbf{a)} Cutting the skeleton (yellow arrow) and connecting segments (green arrow). \textbf{b)} Steering angle between two segments at an intersection $sA=\alpha_1+\alpha_2$. \textbf{c)} Mask reconstruction by estimating the body width.}
	\label{fig:untangle}
\end{figure}

The skeletons of individuals are computed with Algorithm~\ref{alg:untangle} that consists of two steps: 1) cutting the skeleton (line~2-7) to separate worms fused at the endpoints, and 2) resolving intersections by cutting worms and connecting the cut segments from the same worm (line~10-17): see Figure~\ref{fig:untangle}a. The "matches\_one\_of" operations localize the closest target inside a search area (a search circle with radius 5). 

At an intersection, our heuristic decides to connect segments that form a more or less straight line: It connects the segments with the smallest steering angle $sA$, where $sA$ is defined as the sum of the steering angles between the cut segments and the line that connects the cut segments (Figure~\ref{fig:untangle}b).

\subsection{Reconstructing masks from the skeletons}
\label{ssec:recons}
To reconstruct segmentation masks from the skeletons (Figure~\ref{fig:untangle}c), we need to estimate the width of the worm body at each skeleton point. First, edges are detected in the original image with the Canny edge detector~\cite{canny}. For each skeleton pixel, we use the shortest distance to the edge as the radius and fill a circle centered at the skeleton pixel.

In order to make the reconstruction more stable, the estimated radius is smoothed (two pixels before and after) along the skeleton. In addition, we limit the radius to be less than the path length from the skeleton pixel to the skeleton end, so that the segmentation will form a tip at the endpoint. 

\section{Image data}
\label{sec:dataset}

We evaluated our method on two data sets (Figure~\ref{fig:res}): a motivating potato cyst nematode (\emph{Globodera} spp.) data set recorded by PSA, and a public \emph{C. elegans} reference data set (Broad Bioimage Benchmark Collection: BBBC010~\cite{antimicrobial}).
%https://data.broadinstitute.org/bbbc/BBBC010/

The PSA data set contains 3376 nematodes in 1973 microscopic images. Images without nematodes account for 43.6\% in the data set, 26.6\% contain only one, 16.8\% two objects. %Since the samples are obtained from soil by the MEKU-Soil sample extractor (MEKU Erich Polläne GmbH, Wennigsen, Germany), which combines elutriation and sieving, there are still many disturbing objects in the final extractions. 
Since cyst nematodes like \emph{Globodera} spp. produce their offspring in a cyst, which needs to be physically crushed to release worm shaped nematode stages, the samples contain large numbers of distracting objects, such as cysts wall fragments and cyst attached organic material.    
In contrast, the images of the BBBC010 data set contain only nematodes, such that segmentation is easy. However, overlap occurs more frequently due to the worm density that is higher than for PSA.

\section{Results and discussion}
\label{sec:results}

\begin{figure}[h]
	\begin{minipage}[b]{.325\linewidth}
		\centering
		\centerline{\includegraphics[width=\linewidth]{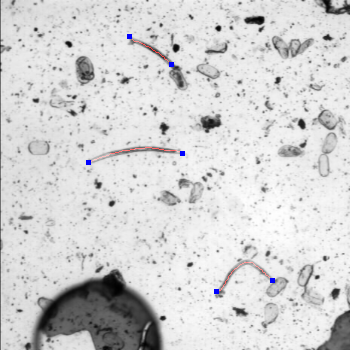}}
	\end{minipage}
	\hfill
	\begin{minipage}[b]{0.325\linewidth}
		\centering
		\centerline{\includegraphics[width=\linewidth]{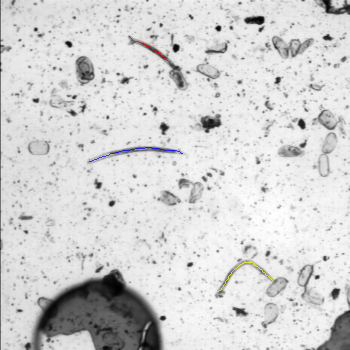}}
	\end{minipage}
	\hfill
	\begin{minipage}[b]{0.325\linewidth}
		\centering
		\centerline{\includegraphics[width=\linewidth]{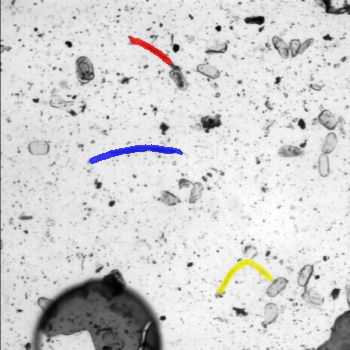}}
	\end{minipage}

	\vspace{0.08cm}

	\begin{minipage}[b]{.325\linewidth}
		\centering
		\centerline{\includegraphics[width=\linewidth]{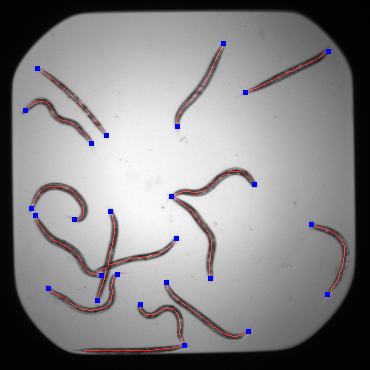}}
		
%		\centerline{(b) Results 3}\medskip
	\end{minipage}
	\hfill
	\begin{minipage}[b]{0.325\linewidth}
		\centering
		\centerline{\includegraphics[width=\linewidth]{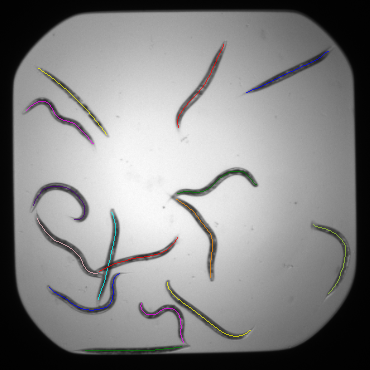}}
		%  \vspace{1.5cm}
%		\centerline{(c) Result 4}\medskip
	\end{minipage}
	\hfill
	\begin{minipage}[b]{0.325\linewidth}
		\centering
		\centerline{\includegraphics[width=\linewidth]{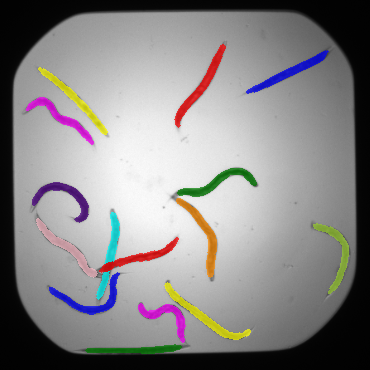}}
		%  \vspace{1.5cm}
%		\centerline{(c) Result 4}\medskip
	\end{minipage}
	\caption{Qualitative results for PSA (top row) and BBBC010 (bottom row). From left to right: predicted skeleton and endpoints, untangled individual worm skeletons and reconstructed segmentation masks.}
	\label{fig:res}
\end{figure}

Figure~\ref{fig:res} shows exemplary qualitative results demonstrating that our approach works robustly on samples with and without distractors, as well as for dense and sparse object collections.

%\begin{figure}[h]
%	\begin{minipage}[b]{0.49\linewidth}
%		\centering
%		\centerline{\includegraphics[width=\linewidth]{figs/psa_curve_ep.png}}
%	\end{minipage}
%    \hfill
%    \begin{minipage}[b]{.49\linewidth}
%    	\centering
%    	\centerline{\includegraphics[width=\linewidth]{figs/bbbc_curve_ep.png}}
%    	%  \vspace{1.5cm}
%    	%		\centerline{(b) Results 3}\medskip
%    \end{minipage}
%	
%	\begin{minipage}[b]{0.49\linewidth}
%		\centering
%		\centerline{\includegraphics[width=\linewidth]{figs/psa_individual_curve.png}}
%	\end{minipage}
%    \hfill
%    \begin{minipage}[b]{0.49\linewidth}
%    	\centering
%    	\centerline{\includegraphics[width=\linewidth]{figs/bbbc_individual_curve.png}}
%    	%  \vspace{1.5cm}
%    	%		\centerline{(c) Result 4}\medskip
%    \end{minipage}
%    
%	\begin{minipage}[b]{0.49\linewidth}
%		\centering
%		\centerline{\includegraphics[width=\linewidth]{figs/psa_individual_mask.png}}
%	\end{minipage}
%    \hfill
%	\begin{minipage}[b]{0.49\linewidth}
%		\centering
%		\centerline{\includegraphics[width=\linewidth]{figs/bbbc_individual_mask.png}}
%		%  \vspace{1.5cm}
%		%		\centerline{(c) Result 4}\medskip
%	\end{minipage}
%	%
%	\caption{Qualitative results for PSA (left column) and BBBC010 (right column). From top to bottom: predicted skeleton and endpoints, untangled individual worm skeletons and reconstructed segmentation masks.}
%	\label{fig:res}
%	%
%\end{figure}

\subsection{Evaluation metric}
\label{ssec:metric}

We quantified object detection performance for individual worms by computing F-scores. This measure is similar to the IoU commonly used in object detection, and it allows us to compare results on BBBC010 with those reported in~\cite{wormToolBox}. We measured precision and recall for different F-score thresholds above which a worm detection was considered to be correct.
%IoU (intersection over union) is also widely used in the literature, which essentially reflects similar standards as F-factor. And we chose F-factor for comparison reason (used by ~\cite{wormToolBox}).

We evaluate both skeletons and masks in this work. When evaluating skeletons (single-pixel), small deviations should be allowed. We formulate the calculation of overlap between prediction and the ground truth as a maximum bipartite matching problem, with each predicted skeleton pixel connected to ground truth pixels within a range of 3.    

\subsection{Experiments on the PSA data set}

The PSA data set contains line and mask annotations as ground truth. It was collected on 10 different dates, and we chose the 367 images from the last three dates as test and the others as training data. To avoid training instability, we ignored images without worms during training.

Our approach achieved 90.34\% precision and 86.28\% recall for mask detection at a F-score threshold of 0.5. After inreasing the F-score threshold to 0.8, the precision was still 75.85\%, with 73.02\% of the worms detected (Table~\ref{sses:res_bbbc}). 

It is worth noting that the performance drop from skeleton to mask was more pronounced on the PSA data set, for example $\approx7\%$ at a F-score threshold of 0.8, while no significant decline could be observed on BBBC010. This resulted from the smaller worm widths in the PSA data set for which the error of width estimation (Section~\ref{ssec:recons}) was relatively larger. 

Analyzing the effect of balanced labels (loss L, Section~\ref{ssec:skel}), we trained a standard segmentation U-Net with unbalanced labels using a binary cross-entropy loss: The model turned out to predict all pixels as background, the larger class (data not shown).

Finally, studying the effect of the ground truth slack (Section~\ref{ssec:skel}), we trained the model using Gaussians with different standard deviations. Table~\ref{tab:perfo} shows that the variants with ground truth slack performed cleary better than those without.

\label{sses:res_psa}
\begin{table*}[h]
	\begin{center}
		\begin{tabular}{c|c|ccccc}
			\hline
			\multicolumn{2}{c|}{\diagbox{Pre./Rec. (\%)}{F-score}} & 0.5 & 0.6 & 0.7 & 0.8 & 0.9 \\ 
			\hline
			\multirow{3}{*}{PSA\_skeleton} & no slack & 89.67 / 84.35 & 76.93 / 82.24 & 56.56 / 70.71 & 42.51 / 53.76 & 28.65 / 36.24 \\
			\cline{2-7}
			& slack\_2\_3 & 90.38 / 88.00 & \textbf{89.55 / 87.41} & 85.75 / 84.82 & 81.00 / 80.24 & 71.14 / 70.47 \\
			\cline{2-7}
			& slack\_3\_5 & \textbf{90.94 / 87.29} & \textbf{89.86 / 86.82} & \textbf{87.44 / 85.18} & \textbf{82.73 / 80.59} & \textbf{75.24 / 73.29} \\
			\hline
			
			\multirow{3}{*}{PSA\_mask} & no slack & 85.49 / 82.91 & 64.19 / 76.16 & 47.91 / 59.88 & 31.91 / 39.88 & 4.00 / 5.00 \\
			\cline{2-7}
			& slack\_2\_3 & \textbf{89.90 / 86.98} & \textbf{86.82 / 85.00} & 82.19 / 80.47 & 72.45 / 70.93 & \textbf{9.74 / 9.53} \\
			\cline{2-7}
			& slack\_3\_5 & \textbf{90.34 / 86.28} & \textbf{87.92 / 84.65} & \textbf{83.70 / 80.58} & \textbf{75.85 / 73.02} & 9.66 / 9.30 \\
			\hline
			
			\multirow{3}{*}{BBBC010\_skeleton} & no slack & 97.68 / 94.51 & 85.89 / 93.98 & 62.68 / 84.30 & 47.35 / 65.47 & 37.32 / 51.60 \\
			\cline{2-7}
			& slack\_2\_3 & 96.89 / 94.66 & 94.43 / 94.51 & 89.08 / 92.99 & 82.72 / 87.20 & 78.02 / 82.24 \\
			\cline{2-7}
			& slack\_3\_5 & \textbf{97.82 / 95.12} & \textbf{95.07 / 94.59} & \textbf{89.91 / 93.22} & \textbf{83.02 / 87.20} & \textbf{78.45 / 82.39} \\
			\hline
			
			\multirow{3}{*}{BBBC010\_mask} & no slack & 87.86 / 96.44 & 72.32 / 89.83 & 59.20 / 76.67 & 46.51 / 60.24 & 19.44 / 25.18 \\
			\cline{2-7}
			& slack\_2\_3 & \textbf{95.94 / 97.08} & \textbf{93.92 / 95.52} & \textbf{90.91 / 92.46} & \textbf{84.20 / 85.63} & \textbf{35.03 / 35.63} \\
			\cline{2-7}
			& slack\_3\_5 & \textbf{96.01 / 96.80} & \textbf{94.12 / 95.45} & 90.34 / 91.75 & 82.91 / 84.21 & 27.73 / 28.17 \\
			\hline
		\end{tabular}
		\caption{Detection precision / recall at different F-score thresholds for skeleton/mask detection on the PSA/BBBC010 data sets. Ground truth slack (Section~\ref{ssec:skel}) variants: slack\_2\_3 refers to the skeleton slack with the standard deviation of the Gaussian $\sigma=2$ and the endpoint slack with $\sigma=3$.}
		 %Bold numbers highlight the best detection precision and recall among the different slack variants.
		\label{tab:perfo}
	\end{center}
\end{table*}
\subsection{Experiments on the BBBC010 data set}
\label{sses:res_bbbc}

For BBBC010, object masks are provided as ground truth. We hence used the morphological skeleton of the ground truth masks to train our model with line annotations. 

In addition, we employed data augmentation to increase the size of the training set: We performed gamma correction with $\gamma=0.5, 2$ to generate images with different contrast. Afterwards, original and generated images were rotated with a step size of 30 degrees. Overall, the training set was expanded to 36 times its original size. 

For training, we split the BBBC010 data set into two parts (A01-B24, C01-E04) and performed two-fold cross validation. Our approach achieved 84.20\% precision and 85.63\% recall for masks at a F-score threshold of 0.8 (Table~\ref{sses:res_bbbc}). As for PSA, we found that the variants with ground truth slack performed better than those without (Table~\ref{sses:res_bbbc}).

On the clean background data BBBC010, also the segmentation/untangling method from~\cite{wormToolBox} can be applied. Our results are superior to the "81\% correctly segmented worms" (mask recall) reported at the 0.8 threshold in~\cite{wormToolBox}.
\section{Conclusions}
\label{sec:conclusion}

We have proposed a CNN-based framework for detecting worm-shaped objects. With the focal loss and the ground truth slack strategy, the CNN model can predict worm skeletons and endpoints robustly. Individuals are untangled from worm clusters, and finally we reconstruct a segmentation mask based on the skeleton. The overall pipeline requires only line (skeleton) annotations for training and outputs segmentation masks. Employing a CNN enables our framework to cope also with images with complex background as they occur in the PSA data set. Future work will focus on improving segmentation accuracy, as well as on extracting descriptive features for nematode phenotyping.                                      

%\vfill
%\pagebreak

\bibliographystyle{IEEEbib}
\bibliography{strings,refs}

\end{document}